\documentclass{article}
\usepackage[english]{babel}
\usepackage{times}
\usepackage[letterpaper,top=2cm,bottom=2cm,left=3cm,right=3cm,marginparwidth=1.75cm]{geometry}
\usepackage{soul}
\usepackage{xcolor} 
\usepackage{amsmath}
\usepackage{amsfonts}  
\usepackage{graphicx}
\usepackage{subcaption}
\usepackage[colorlinks=true, allcolors=blue]{hyperref}
\usepackage{algorithm}
\usepackage{algpseudocode}
\usepackage{booktabs} 
\usepackage{multirow}
\usepackage{makecell}
\usepackage{array} 
\usepackage{mathtools}
\usepackage[normalem]{ulem}

\title{AI-Based Simulator for 3D Indoor Map Navigation and Pathfinding }

\author{
Ali Bayeh\textsuperscript{1}
\thanks{ORCID: \href{https://orcid.org/0009-0006-6402-9402}{0009-0006-6402-9402}}
\quad\quad
Samira Sadaoui\textsuperscript{1}
\thanks{ORCID: \href{https://orcid.org/0000-0002-9887-1570}{0000-0002-9887-1570}}
\quad\quad
Saurabh Kapoor\textsuperscript{2}
\\[1ex]
\textsuperscript{1}Department of Computer Science, University of Regina, SK, Canada\\
\textsuperscript{2}Prisau Innovation Limited, Regina, SK, Canada \\
{\{alibayeh, sadaouis\}@uregina.ca; im.saurabhkapoor@gmail.com}}

\date{}

\begin{document}
\maketitle
\begin{abstract}
Our study develops an AI-based simulator that transforms static indoor maps into interactive environments for navigation and wayfinding. The simulator integrates image preprocessing, semantic image segmentation, 2D and 3D reconstruction, pathfinding and interactive visualization. To address the scarcity and incompleteness of labeled real-world indoor plan datasets, we produce our own training dataset by constructing diverse structural map elements. Our segmentation model is built on a SegFormer backbone, enhanced with a deformable CNN stem, Atrous Spatial Pyramid Pooling fusion, attention-based skip connections, multiple prediction heads, and inference-time refinement. For 3D reconstruction, the predicted segmentation maps are converted into polygonal geometries and subsequently into 3D meshes, enabling interactive exploration. The simulator provides four viewing modes: Orbit, Walk, Auto-Walk and  2D Minimap. The Orbit View allows users to explore maps in both 2D and 3D through interactive rotation and zooming. The Walk View enables users to navigate and explore the 3D environment. The Auto-Walk View provides an animated walkthrough of the calculated shortest path route from the starting point to the destination. Our web-based tool aims to enhance the user experience by facilitating pathfinding, improving accessibility and supporting scenario-based training in large, complex and unfamiliar indoor environments.
\end{abstract}


\section{Introduction}
\subsection{Research Scope}
Static maps are commonly used to represent layouts of indoor spaces, but they can be difficult to understand and analyze as they may contain many spatial elements and symbols and several possible routes to reach the destination. Moreover, they provide no support for an efficient wayfinding, accessibility-based exploration and scenario-based training, for example, for emergency evacuation. Our main goal is to develop an efficient and user-friendly AI-based simulator that allows users explore and test complex and unfamiliar indoor spaces, such as hospitals, office buildings, airports, universities, schools, stadiums, museums, parking lots, commercial malls and evacuation maps. Our simulator transforms indoor map images into interactive experiences that help users become familiar with an indoor space before exploring it in real life, and also provides the shortest route between two user-specified locations. Our tool offers an engaging way to explore valid routes and practice decision-making before entering unknown spaces.    

\subsection{Contributions}
Our paper presents a practical AI-driven simulator to simplify wayfinding by transforming static maps into interactive and actionable navigation systems. It reduces confusion, stress and the risk of getting lost in unfamiliar and large spaces, improves accessibility by accommodating users with diverse mobility needs, and supports safe navigation during evacuation situations. 

The proposed solution is a standalone software system based on advanced deep learning methods, operating independently of third-party AI tools. In addition, we construct our own training dataset using heuristic methods. This independence enables greater flexibility in software customization and supports continuous improvements tailored to the requirements of specific indoor spaces. Our tool consists of indoor-plan image uploading, semantic image segmentation, automated 2D and 3D reconstruction, 3D interactive navigation and pathfinding between two selected locations. We first develop a heuristic data generator to create consistently labeled training samples based on multiple  structural and spatial map elements. Next, a segmentation model predicts a per-pixel mask representing various map elements. Our segmentation architecture FloorFormer extends the SegFormer backbone \cite{SegFormer2021} with a deformable CNN stem, Atrous Spatial Pyramid Pooling (ASPP) fusion, attention-based skip connections, multiple prediction heads and inference-time refinement. Subsequently, geometry processing converts this prediction into a 2D and 3D scenes and meshes for interactive exploration.  

Our simulator supports several view modes: Orbit, Walk, Auto-Walk and Minimap, alongside the navigation route. The application’s Orbit View provides users with an overview of the map in both 2D and 3D, offering interactive features such as map rotation and zooming. The Walk View offers a first-person perspective, similar to that of a video game, allowing users to navigate and explore the 3D environment. The application also features an Auto-Walk View, which provides an animated walkthrough of the calculated route. The navigation algorithm determines the shortest path between the specified start and end points, and the view automatically follows the route from beginning to end. 

The paper is structured as follows. Section 2 is an overview of the five main phases of our AI-based simulator. Section 3 describes in detail the construction of our training indoor-plan dataset based on numerous spatial elements, such as room layout, wall, window, door, furniture, fixture, stairs, elevators and column. Section 4 exposes our indoor image segmentation architecture and underlying components. Section 5 presents our 2D and 3D reconstruction. Section 6 explains the pathfinding and navigation in the 3D representation. Section 7 describes training of the different models. Section 8 presents several examples of our GUI-based simulator. Section 9 concludes our research work and highlights some future work. 
\section{AI-based Simulator: An Overview }
We develop a new methodology to build the AI-based simulator, comprising the following phases:
\begin{itemize}
\item \textbf{Synthetic indoor-plan dataset generation:}
We construct a robust synthetic floor-plan dataset to support the training and evaluation of the AI models. We devise our dataset to capture diverse structural map elements, such as rooms, corridors, exits, stairs and other relevant elements. Introducing variations in floor-plan geometry and layout helps improve the model's ability to generalize to previously unseen environments.

\item \textbf{Indoor-plan image upload:}
The process begins by uploading a digital image of the floor-plan layout into the Web-based tool. The input may represent, for example, an evacuation map, laboratory floor plan, hospital layout, commercial center, or university campus. Users can upload floor-plan images in common formats, such as JPEG, PNG or GIF. This flexibility enables the simulator to process a wide range of two-dimensional maps without requiring specialized input formats.

\item \textbf{Semantic segmentation and map-element prediction:}
The uploaded floor-plan image is first processed using a semantic segmentation model to identify and classify the different map elements at the pixel level. Each pixel is assigned to a corresponding semantic class, such as room, corridor, door, entrance and stair. This stage transforms the original image into a structured representation that can be used for subsequent spatial analysis and navigation.

\item \textbf{2D and 3D reconstruction and interactive visualization:}
The segmented and predicted floor-plan representation is subsequently reconstructed into a 2D and 3D models. The 3D visualization provides users with an interactive representation of the environment, allowing them to explore the reconstructed space and better understand its spatial organization. This stage also provides the visual foundation for interaction and route-planning functionality.

\item \textbf{Route pathfiding and navigation:}
Users can select start and destination points directly on the interactive map (2D and 3D). Before route generation, the simulator verifies that the selected locations are positioned within accessible areas and at least one valid pathway exists between them. We employ the known pathfinding algorithm A*~\cite{AstarAlg1968} to compute a feasible route between the selected points. This method operates on the reconstructed navigable representation of the floor plan while avoiding non-traversable elements, such as walls and other obstacles. If multiple feasible routes exist, the method can further identify an efficient route according to criteria, such as travel distance or path cost. This functionality enables a robust route planning and navigation within the reconstructed environment using different exploration views.

\end{itemize}
\section{Generation of Synthetic Indoor-Plan Dataset}

\begin{figure}[h!]
\centering
    \begin{minipage}[b]{0.48\textwidth}
        \centering \includegraphics[ width=\linewidth]{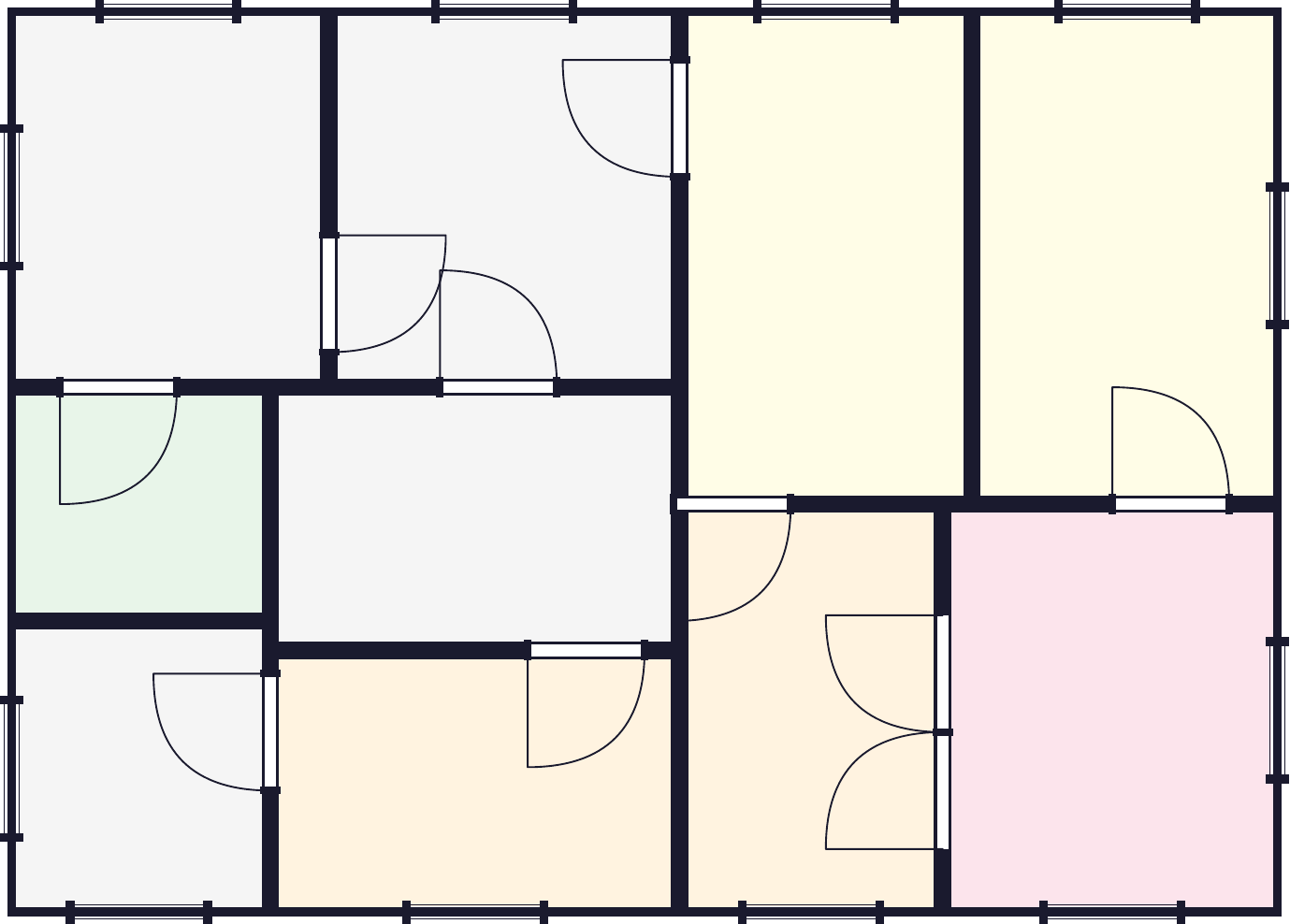}
        \par
        \vspace{-0.5em}
        \caption*{(a) }
    \end{minipage}
    \hspace{1em}
    \begin{minipage}[b]{0.48\textwidth}
        \centering        \includegraphics[ width=\linewidth]{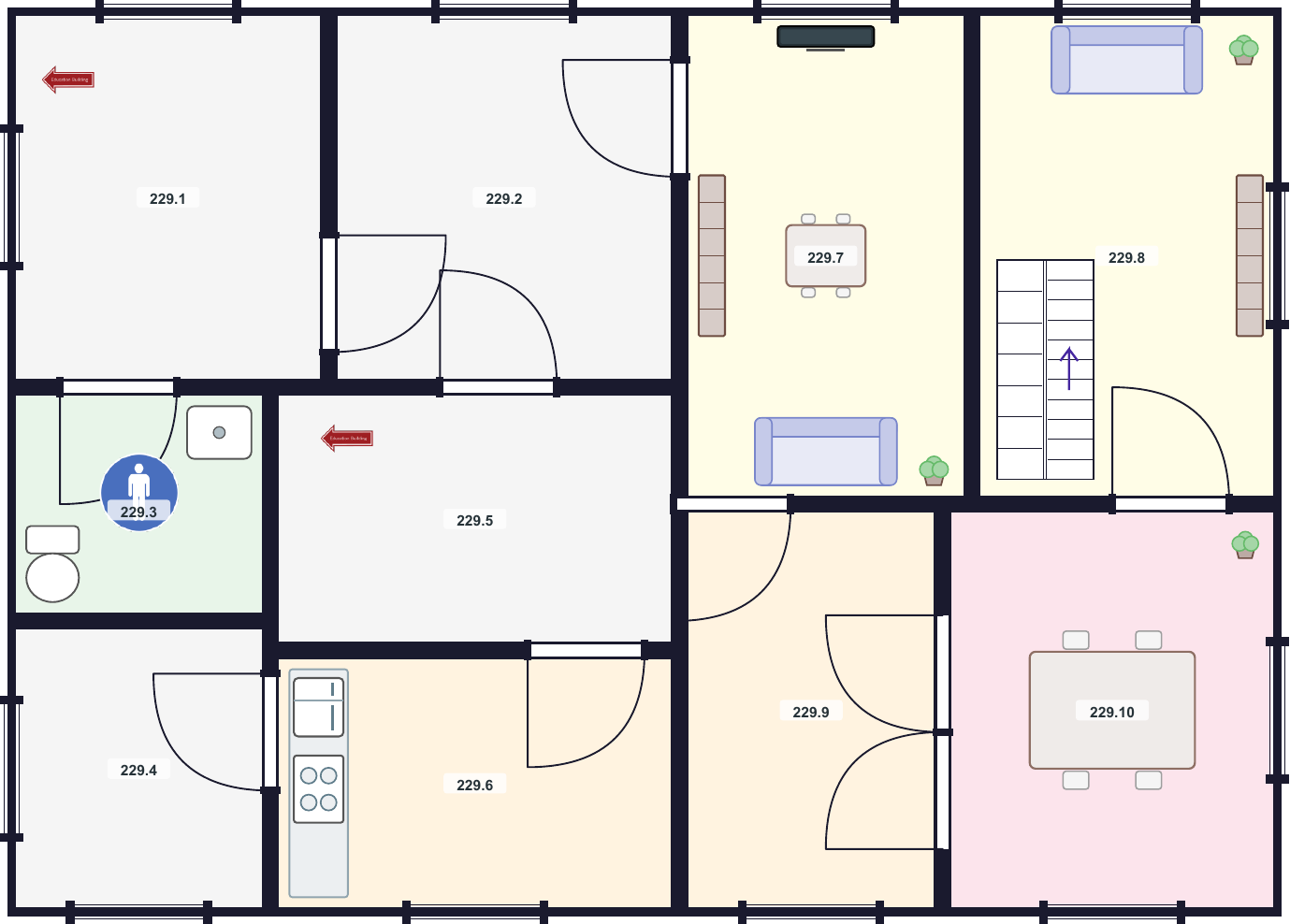}
        \par
        \vspace{-0.5em}
        \caption*{(b) }
    \end{minipage}

    \caption{First generation of floor plans with/without details: (a) The algorithm begins with a simple rectangle, using the binary space partitioning method to construct a random-size space (with predefined min/max size) and assign a label to each partition, (b) Based on the assigned labels, the algorithm adds details such as furniture, as well as guide symbols, to random locations within each partition.}
    \label{fig:first_generation}
\end{figure}

Because labeled real floor-plan datasets are scarce and incomplete, we generate the training dataset synthetically. Instead of training a deep generative model, we develop a robust and efficient heuristic pipeline to construct a plan from a base template.
Our heuristic algorithm essentially acts as a geometric validity filter rather than an architectural simulator and constructs floor plans layer by layer. The known geometry of each layer provides the ground truth for the segmentation mask.

Next, the algorithm treats each floor plan element \footnote{Plan elements such as room polygons, walls, doors, windows, furnitures and guide symbols} as a labeled primitive that must be placed without illegal overlap and within a safe area (with a minimum clearance from corners or other elements). A greedy overlap check then accepts or rejects placements sequentially, so the generator only guarantees that the drawing is renderable and that each pixel has a consistent ground-truth label.
The objective of segmentation is to identify the pixel-wise or structural label map, which remains valid while the mask is generated from the same geometry that produced the image, even when the layout is functionally absurd.
Its noteworthy that the semantic plausibility and segmentation correctness are different requirements. A segmentation model does not need to learn building codes, circulation rules or furniture placement, but it needs to learn boundaries, regions, and class-consistent visual patterns.
Extra doors, blocked doors, or odd window placement may be unrealistic, but they still provide valid examples of walls, openings, room interiors and object masks. The greedy overlap and safe-distance heuristics prevent multiple objects (e.g. furniture items) from being assigned to the same region or an overlapping region. This reduces label collisions, which can obscure symbols, and enforces a minimum separation between objects and doorways. This prevents doors from being placed too close to corners. Overall, these constraints enhance the legibility of floor plan symbols and improve the quality of the generated segmentation annotations overall.
Enforcing real floor-plan semantics would result in a reduction of the search space, an increase in rejection rates, and a bias of the dataset toward narrow conventions, unless a robust generative model is employed. Below, we explain how the different spatial elements are identified and re-constructed:\\

\noindent \textbf{Room Layout:} this layout can be generated using three related spatial-partitioning strategies: 
\begin{itemize}
\item (a) binary space partitioning (BSP), which recursively subdivides a rectangular footprint using horizontal or vertical cuts until each subregion satisfies a specified room-size constraint \cite{KdTrees2010,BSP2019}
\item (b) grid-cell or polyomino-based allocation, which discretizes the footprint into a coarse grid and groups contiguous cells into connected room regions, subject to room-area and adjacency constraints \cite{FlPlanMixCons2021,FlPlanBDD2015}.
\item (c) weighted recursive slicing or treemap-inspired allocation, which partitions the footprint into rectangular regions by selecting cut positions according to target room areas, with a treemap-inspired variant assigning region areas proportionally to prescribed room weights \cite{SlicTree2001,FPlanRoomAreaTreeAdj1988,TreeMap1992}.
\end{itemize}

\noindent \textbf{Walls, Windows and Doors:}  these elements are generated using geometric and connectivity-based rules, as follows:
\begin{itemize}
\item  (a) interior and exterior walls are reconstructed from shared room boundaries and the building footprint, respectively, where polygon offsetting is used to enforce wall thickness and based on the skeleton representation, either the straight skeleton or the medial axis can be used to provide a geometric representation of the polygonal regions \cite{SkelPoly1995}.
\item (b) for door placement, candidate door locations are identified along shared boundaries between adjacent rooms and the room-adjacency graph and geometric constraints are used to ensure valid openings and room connectivity \cite{GenLayHiLevSpec2020,RulSynFlPlan2023}(for minimal connectivity, a spanning tree rooted at the corridor can be extracted using breadth-first search (BFS) or depth-first search (DFS), retaining only the traversal-tree edges as door connections\cite{algorithms2022}).
\item (c) the perimeter-based window placement put Windows only on exterior-facing wall segments, with candidate locations and dimensions constrained by wall geometry and non-overlap requirements \cite{RulSynFlPlan2023}. The number or placement probability of windows may additionally be weighted by the length of each eligible exterior wall segment. This proportional-length rule is adopted as a heuristic to distribute windows according to available wall space.
\end{itemize}

\noindent \textbf{Furniture and Fixtures:}  furniture and other interior objects are placed according to the room labels generated in the preceding stages, using room-specific templates subject to geometric and functional constraints. The placement procedure comprises three components:
\begin{itemize}
\item (a) clearance-constrained placement, in which each furniture item is represented by a two-dimensional footprint (e.g., an oriented bounding box), and candidate placements are rejected if they violate wall boundaries, overlap other objects or obstruct door-swing.

\item (b) greedy placement with conflict-driven backtracking, in which objects are placed sequentially at preferred candidate locations, and previous placements are reconsidered when conflicts prevent subsequent objects from being placed feasibly.

\item (c) room type specific templates, which encode canonical furniture arrangements for different room categories and are adapted to the available room dimensions and the locations of doors and windows. These templates can incorporate functional zones, relative object positions, and ergonomic relationships to produce plausible arrangements.
\end{itemize}

\noindent \textbf{Other Structural Elements:} 
other structural elements, such as stairs, elevators and columns, are added by fitting predefined geometric assets into feasible areas of the generated floor plan as follows: 

(a) deterministic asset fitting helps to ensure that stairs and elevators are assigned to available regions by selecting, orienting and scaling predefined templates to fit the available bounding boxes, all the while respecting geometric constraints.

(b) pillars are positioned using rejection sampling, in which candidate locations are randomly generated and rejected if they violate footprint boundaries, minimum separation distances or collision constraints.\\

\noindent \textbf{Texture and Visual Style Randomization:} to improve robustness to variations in drawing and rendering conventions, we introduce appearance-level domain randomization by varying color palettes, line styles, line widths, and image noise while preserving the underlying floor-plan geometry. This variation reduces systematic correlations between semantic classes and specific visual attributes, encouraging the segmentation model to rely on geometric boundaries and spatial structure rather than fixed color or rendering cues. The strategy follows established principles of domain randomization and image data augmentation, which have been used to improve generalization across visual domains and to encourage shape-based rather than texture-dependent recognition \cite{ImgAug2019,CnnTextImprov2019,TextureBias2020}.\\

\noindent \textbf{Flexibility in Training Data:}
Regarding visual style randomization, while altering the appearance of certain elements may require only minor configuration changes, randomizing core structural elements (such as walls, doors and windows) demands more careful design and consideration. 
For example, real-world floor plans use various symbols to represent doors, such as single-leaf, double-leaf and sliding doors. 
Similar variations exist for walls and windows. Even seemingly simple elements, such as walls, can differ in their graphical representation, line styles and structural details, depending on the software used to produce the floor plan, regional drafting conventions and individual designer preferences. 
While these elements generally retain recognizable geometric characteristics that enable humans to identify them across different styles, a segmentation model may not generalize equally well to all such variations. 
Therefore, to improve generalization and reduce dependence on specific visual representations, synthetic training data should include a sufficiently diverse set of styles for each core element while preserving the underlying geometry and semantic consistency of the annotations.

\begin{figure}[h!]
\centering
    \begin{minipage}[b]{0.25\textwidth}
        \centering \includegraphics[ width=\linewidth]{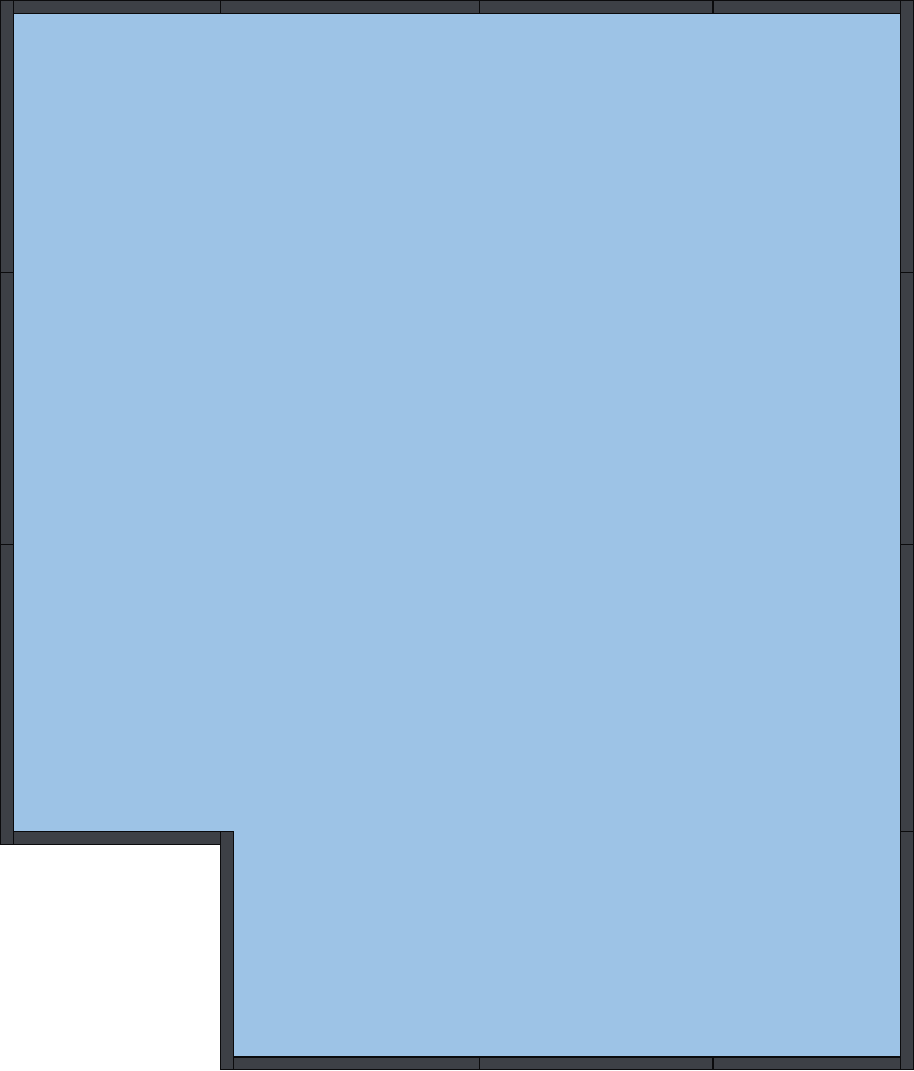}
        \par
        \vspace{-0.5em}
        \caption*{(a) }
    \end{minipage}
    \hspace{8em}
    \begin{minipage}[b]{0.25\textwidth}
        \centering        \includegraphics[ width=\linewidth]{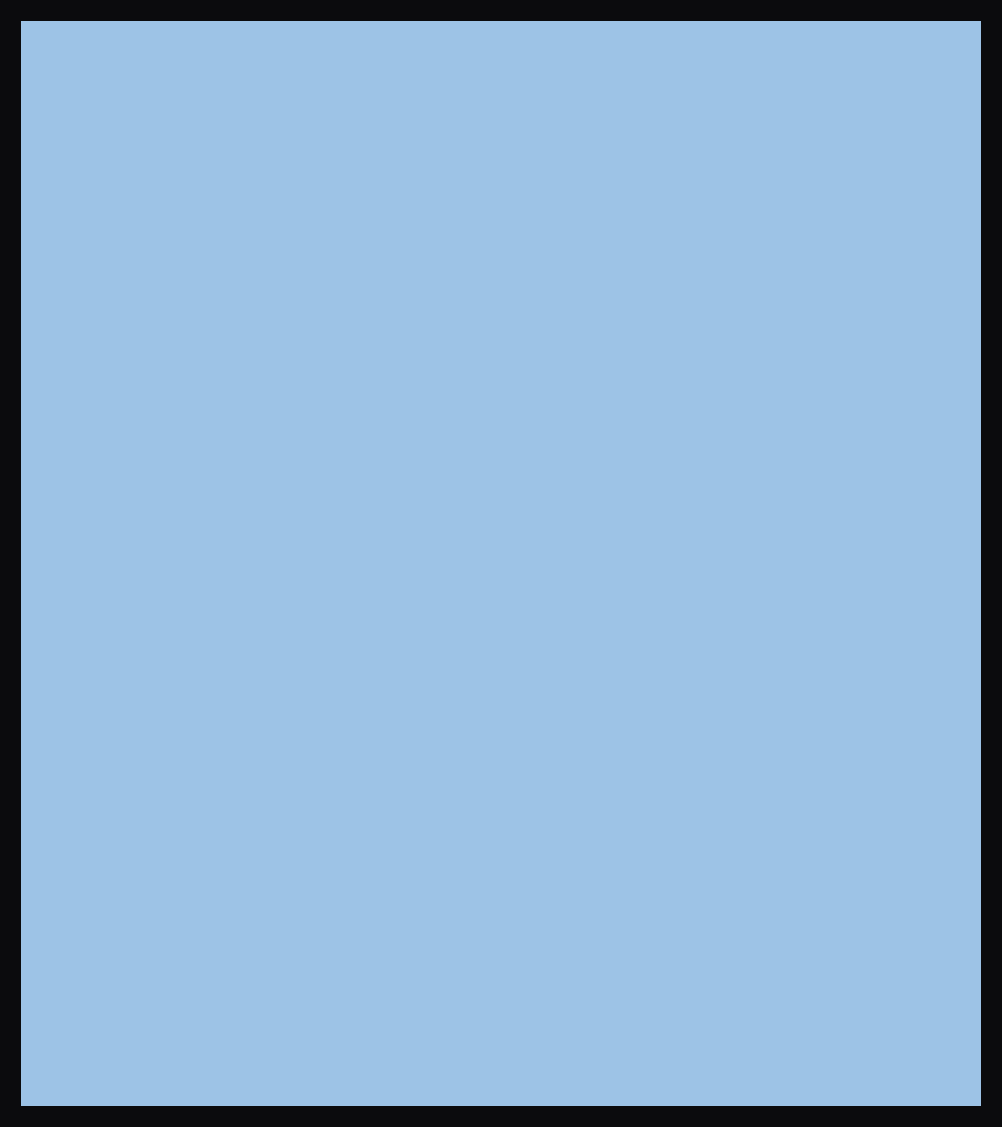}
        \par
        \vspace{-0.5em}
        \caption*{(b) }
    \end{minipage}

    \caption{General layout structure used to generate the dataset}
    \label{fig:general_layout}
\end{figure}

To this end and to make the process more extensible and customizable, the main elements are randomly selected from a list of files, which can be extended by adding a new file for each element. A similar structure is used for the overall layout and other floor plan elements.
Figure \ref{fig:general_layout} shows the general layout structure used to generate the dataset. While the layout's structure remains the same, the number and size of the rooms are vary.

\begin{figure}[h!]
\centering    
    \begin{subfigure}{\textwidth}
    \centering
    \begin{minipage}[b]{0.3\textwidth}
        \centering \includegraphics[ width=\linewidth]{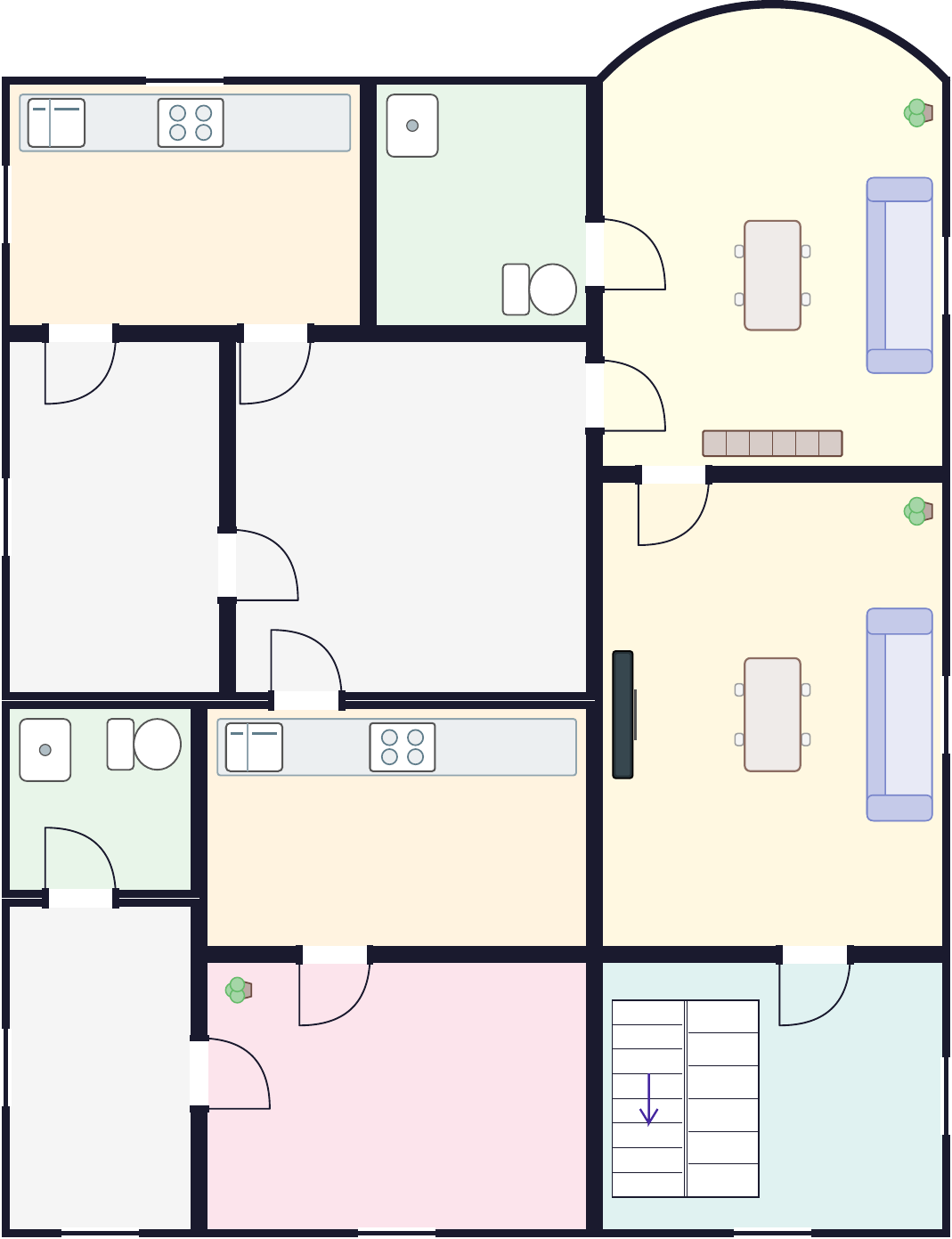}
        \par
        \caption*{(a) A residential sample image with an arc wall}
    \end{minipage}
    \hspace{1em}
    \begin{minipage}[b]{0.3\textwidth}
        \centering        \includegraphics[ width=\linewidth]{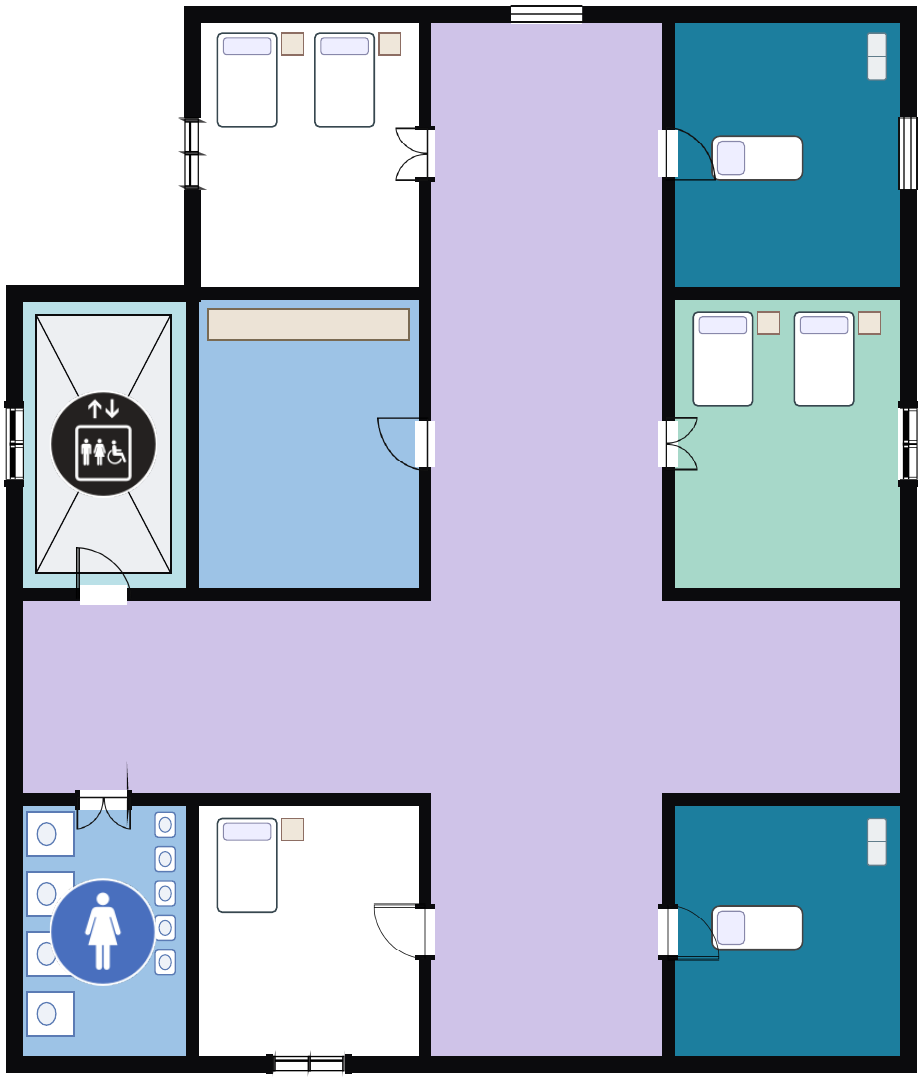}
        \par
        \caption*{(b) A hospital sample image with default layout}
    \end{minipage}
    \hspace{1em}
    \begin{minipage}[b]{0.3\textwidth}
        \centering        \includegraphics[ width=\linewidth]{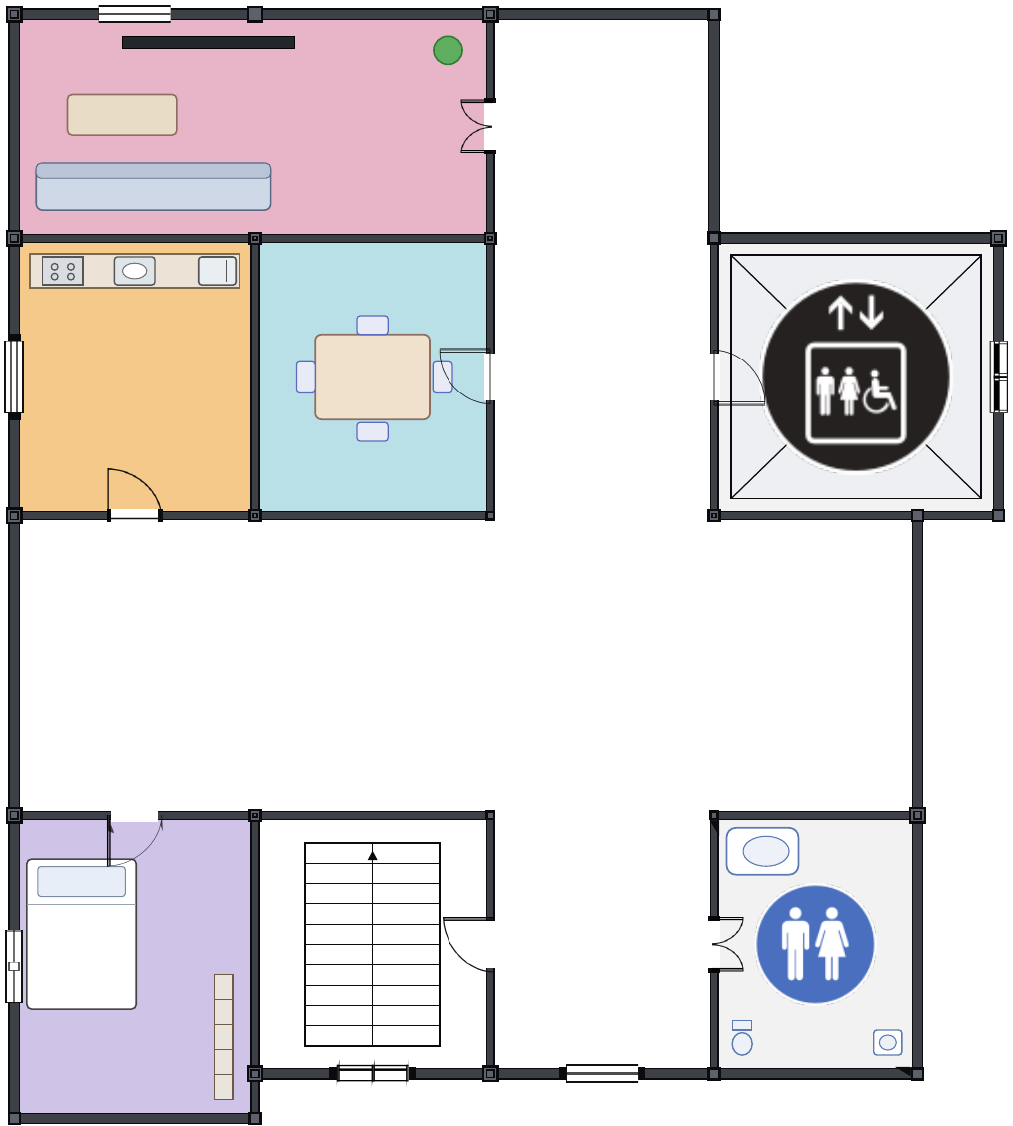}
        \par
        \caption*{(c) A residential sample image with protrusion}
    \end{minipage}
    \end{subfigure}
    
    \vspace{4mm}
    
    \begin{subfigure}{\textwidth}
    \centering
    \begin{minipage}[b]{0.3\textwidth}
        \centering \includegraphics[ width=\linewidth]{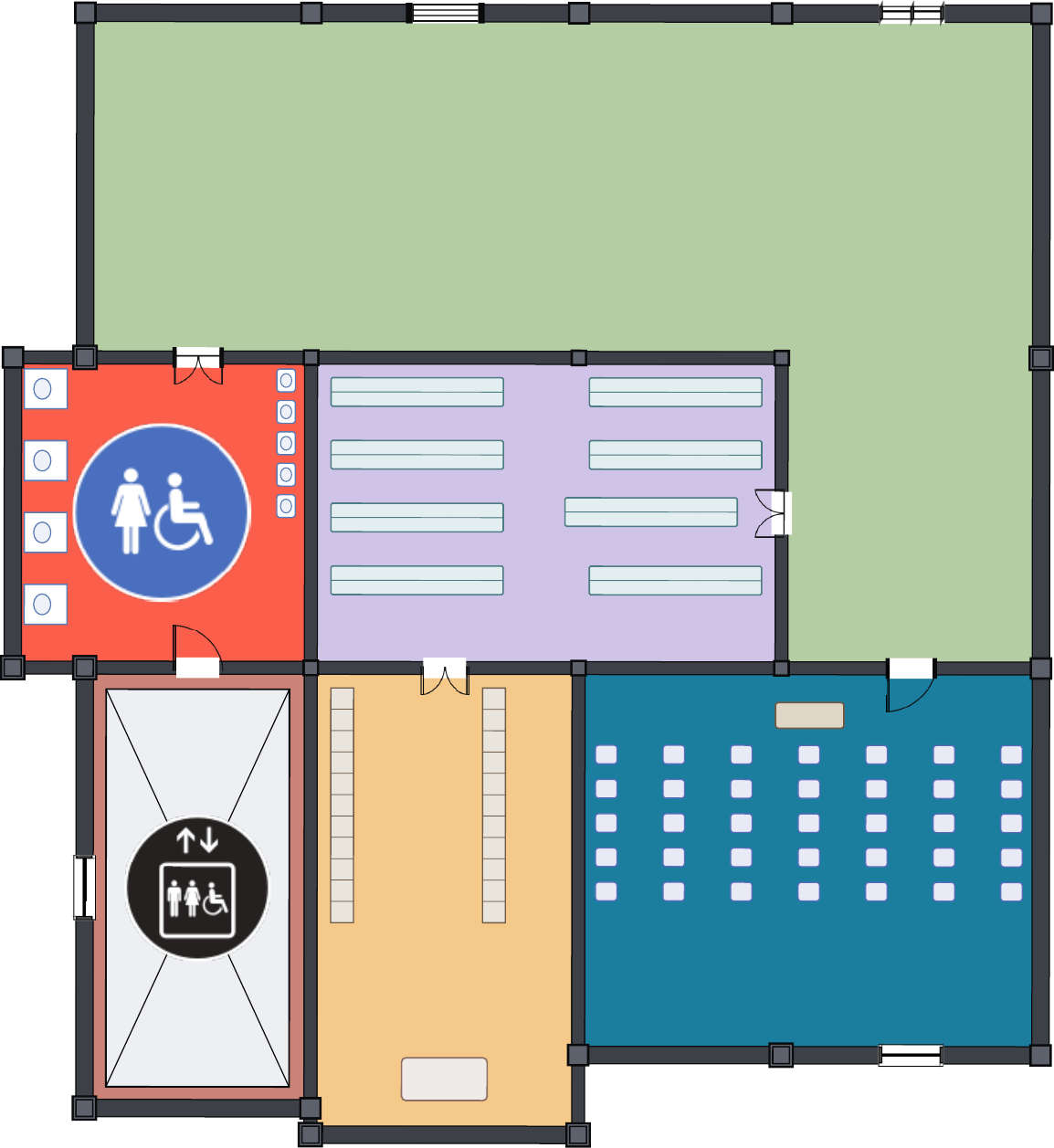}
        \par
        \caption*{(d) An university sample image with protrusion and thick walls}
    \end{minipage}
    \hspace{1em}
    \begin{minipage}[b]{0.3\textwidth}
        \centering        \includegraphics[ width=\linewidth]{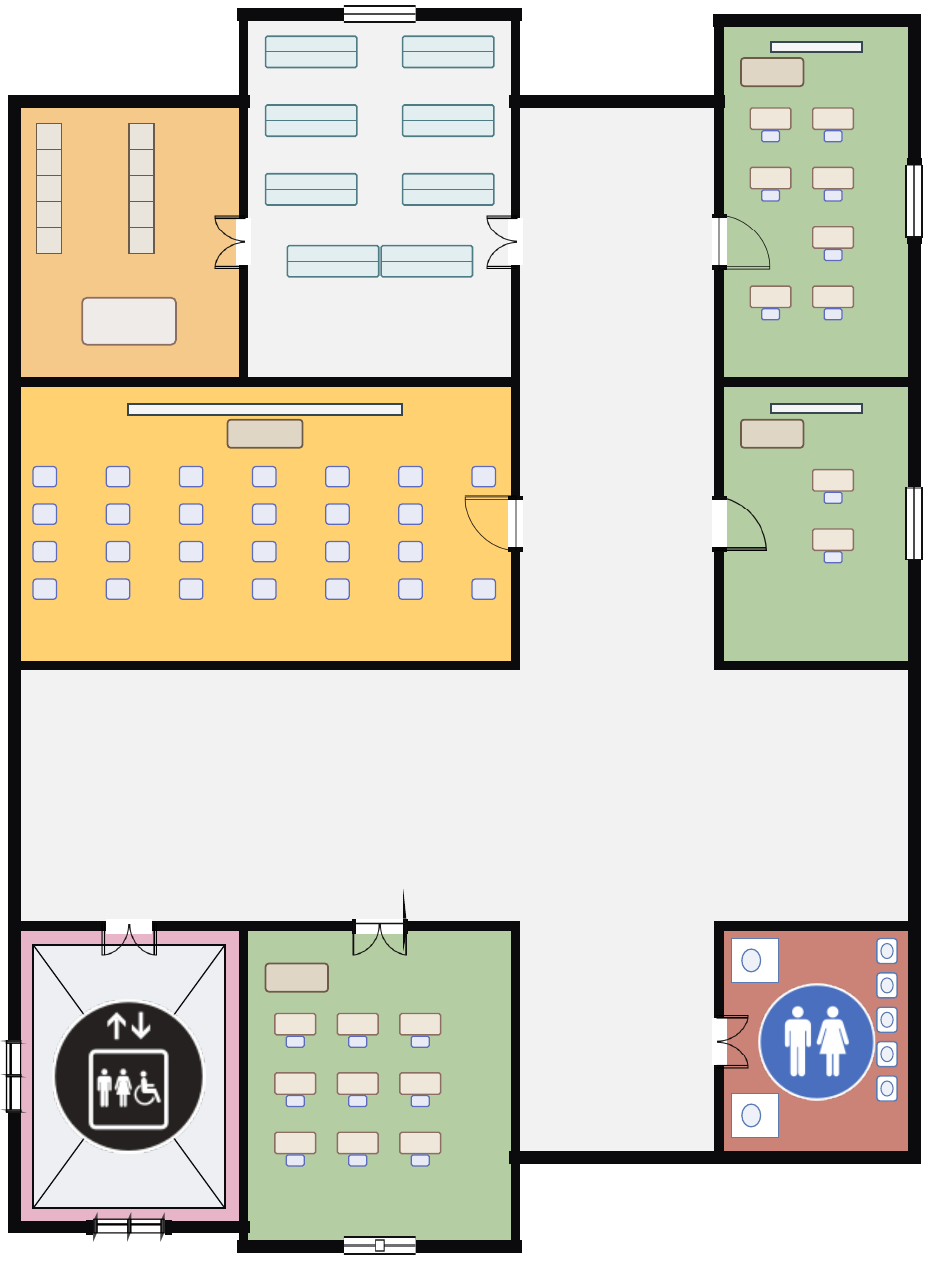}
        \par
        \caption*{(e) An university sample image with protrusion}
    \end{minipage}
    \hspace{1em}
    \begin{minipage}[b]{0.3\textwidth}
        \centering        \includegraphics[ width=\linewidth]{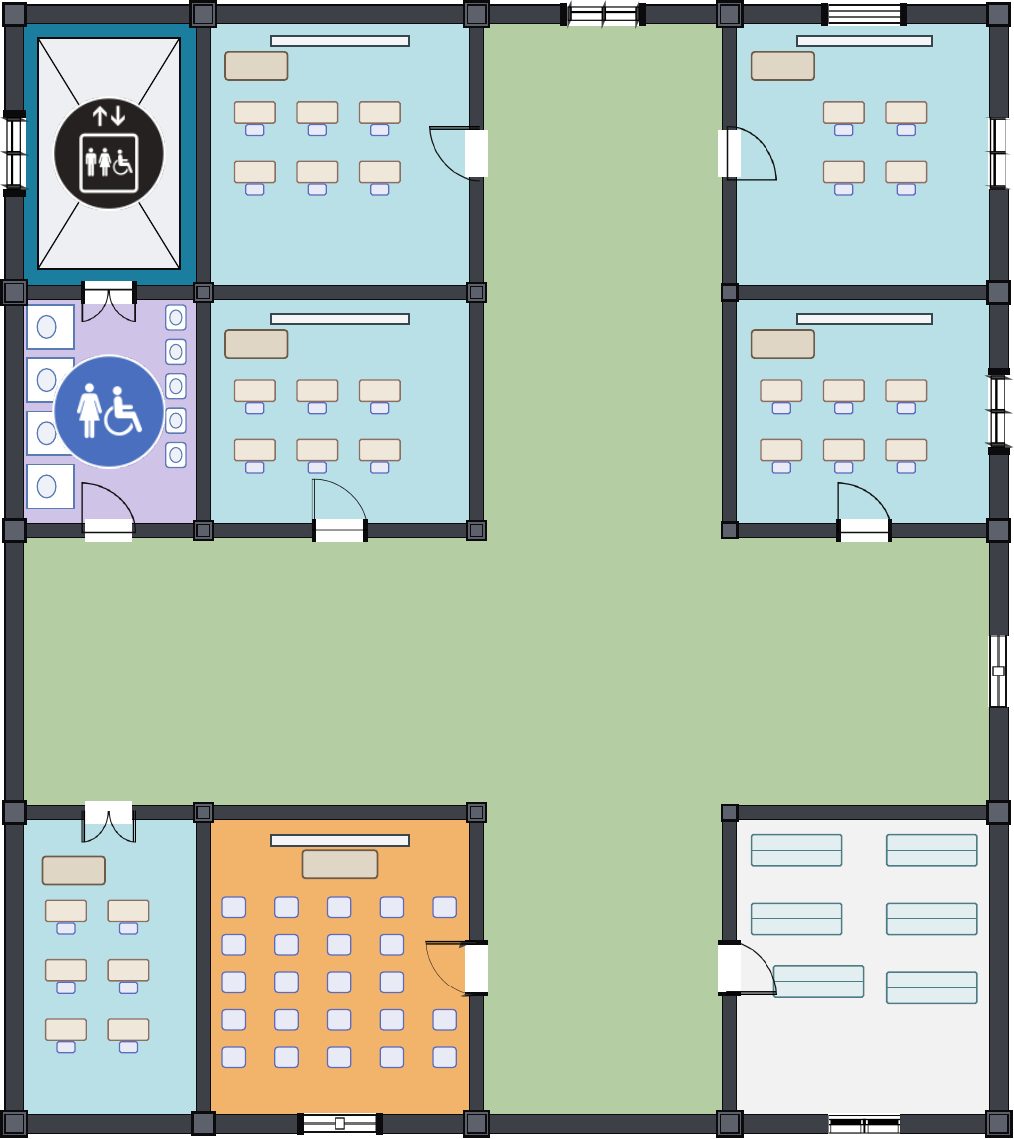}
        \par
        \caption*{(f) An university sample image with default layout and thick walls}
    \end{minipage}
    \end{subfigure}    
    
    \caption{The sample images from each class in the final dataset. To make the model robust against stylistic details, the images vary in terms of color, graphical representation of floor plan elements, and even the width of walls and windows.}
    \label{fig:final_dataset}
\end{figure}

To generate a more diverse dataset, we consider applying additional techniques to each layout after the rooms have been generated. The first is the protrusion technique, where one room is randomly selected and, if certain constraints such as label and location are met, the room will protrude from the outer wall. The second is the arc structure, where one room is randomly selected and, if the location allows, the straight outer wall will be replaced with an arched wall.  
Figure \ref{fig:final_dataset} illustrates examples from the final dataset. We consider three different layout classes: 1) residential, 2) hospital, and 3) university. For each class, we incorporate specific furniture, equipment, and fixtures appropriate to that environment such as TVs, sofas, and plants for residential images; hospital furniture and medical equipment for hospital scenes; and educational and classroom furniture for university settings.

The final dataset consists of $16,000$ generated images, divided into $10,000$ training images, $3,000$ validation images and $3,000$ test images. Different random seeds and hyperparameter configurations were used during image generation to introduce variation across the three subsets. Notably, the training set incorporates a wider range of geometric modifications, such as rotation and protrusion, compared to the validation and test sets. Furthermore, the diversity of architectural symbols varies across the subsets. Specifically, the validation and test sets include a broader variety of door and window symbols than the training set. These variations enable the evaluation of the model's ability to generalize to different geometric configurations and symbol representations.

\section{FloorFormer Image Segmentation Model}
\begin{figure}
    \centering
    \includegraphics[ width=0.92\linewidth]{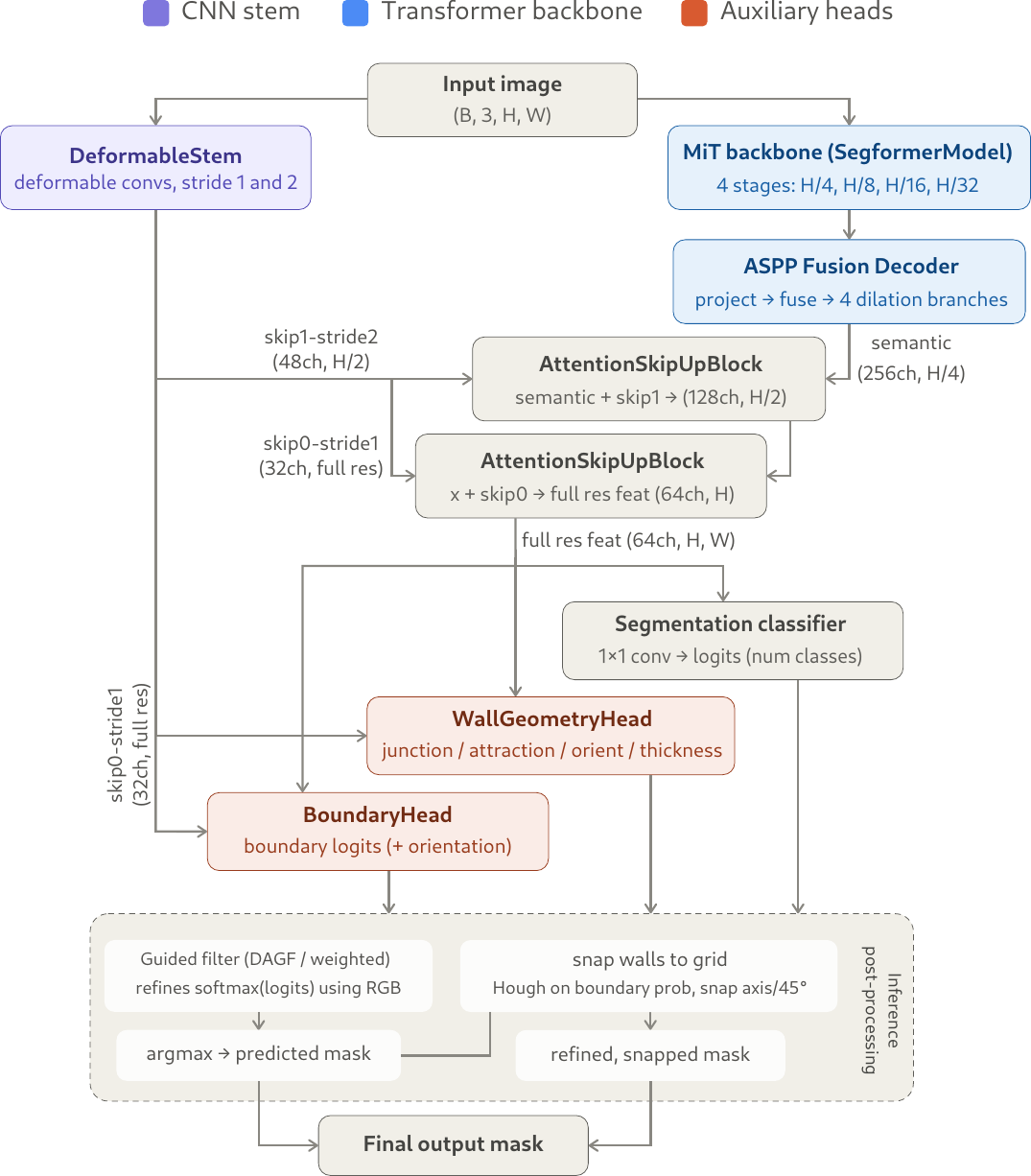}
    \caption{FloorFormer segmentation model architecture. Although the Segformer model is trained as the backbone of the segmentation, due to the image being progressively downsampled by its hierarchical encoder across four distinct stages, the model will miss details of thin objects such as walls. The DeformableStem branch works on the original input image to mitigate the loss of detail in the final output}
    \label{fig:architecture}
\end{figure}
    
The image segmentation model, FloorFormer, consists of the following main components: 
\begin{itemize}
\item \textbf{Dual-path Encoder:} A deformable convolutional stem extracts high-resolution skip features, while a SegFormer MiT backbone \cite{SegFormer2021} produces hierarchical features at progressively lower spatial resolutions. 
\item \textbf{ASPP Fusion and Attention Skip Up-blocks:} Atrous Spatial Pyramid Pooling (ASPP) fuses the backbone features, and attention-based up-sampling blocks restore resolution using the stem features.
\item \textbf{Three Task Heads:} The model contains three output heads: a semantic segmentation classifier, a boundary head with orientation information, and a wall-geometry head for junction, attraction and thickness information. 
\item \textbf{Inference-only Post-processing:} During inference, guided filtering refines the soft mask and Hough-based processing can be used to align wall boundaries. 
\item \textbf{Training Process:} Model training combines cross-entropy and Dice losses with auxiliary deep supervision, boundary and orientation losses, and ramped geometry loss. 
\end{itemize}
\section{3D Reconstruction and Interactive Visualization}
The predicted segmentation mask is then refined through automated post-processing. 
This uses classical image-processing techniques to correct potential labeling errors, improve boundary accuracy and refine the segmentation mask. 
Specifically, contour-based processing and majority voting among neighboring pixels are employed to refine region boundaries and improve label consistency. 
Furthermore, color-thresholding cues extracted from the original input image are used to rectify potential segmentation errors and align the predicted boundaries more closely with the visible structures in the image. Finally, the refined mask is converted into 2D and 3D scenes:
\begin{itemize}
\item Shapely extracts polygons and performs 2D geometry operations on room and wall footprints.
\item Trimesh extrudes these shapes into 3D meshes and assembles the scene, which is exported into glTF format for use in a browser-based viewer for example.
\end{itemize}

\section{Route Pathfinding and Navigation}
For path planning, the semantic segmentation mask generated by the FloorFormer model is converted into a binary traversability map, where wall and window pixels are considered as obstacles, while all remaining regions, including doorways, are considered traversable. 
Then we apply a distance transform to calculate the distance from each traversable pixel to the nearest obstacle pixel. 
Specifically, we employ the L2 distance transform with a "5×5" neighborhood, which provides an approximation of the Euclidean distance using a chamfer-based approach \cite{DistTransfor1986}, as implemented in OpenCV \cite{OpenCV2000}. 
The resulting distance values are rounded to the nearest integer and used to construct a grid-based navigation representation that encodes the clearance available at each traversable cell. 

Our simulator supports three navigation modes: (a) In Auto-Walk, the user selects two points (start and end) and  A* search algorithm finds a shortest route across the generated mesh, after which the camera follows the path in first-person view, (b) Manual-Walk allows the user to move with keyboard and mouse controls, and (c) Orbit mode provides a static overview for orientation. These functions make route learning interactive and game-like. If multiple feasible routes exist, the method can further identify an efficient route according to criteria, such as travel distance or path cost.
\begin{figure}[h]
\centering    
\includegraphics[width=0.8\linewidth]{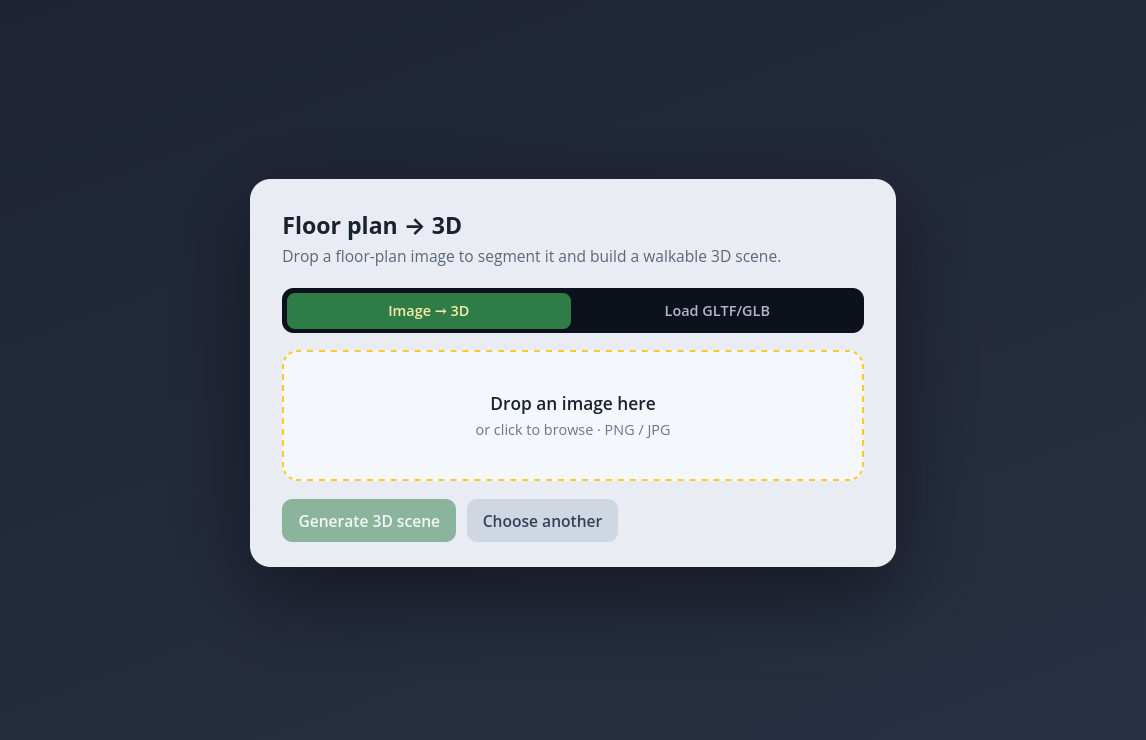}
\caption{Simulator Web Interface - Home Page}
\label{fig:gui-runtime}
\end{figure}

\begin{figure}[h!]
\centering    
\includegraphics[width=0.8\linewidth]{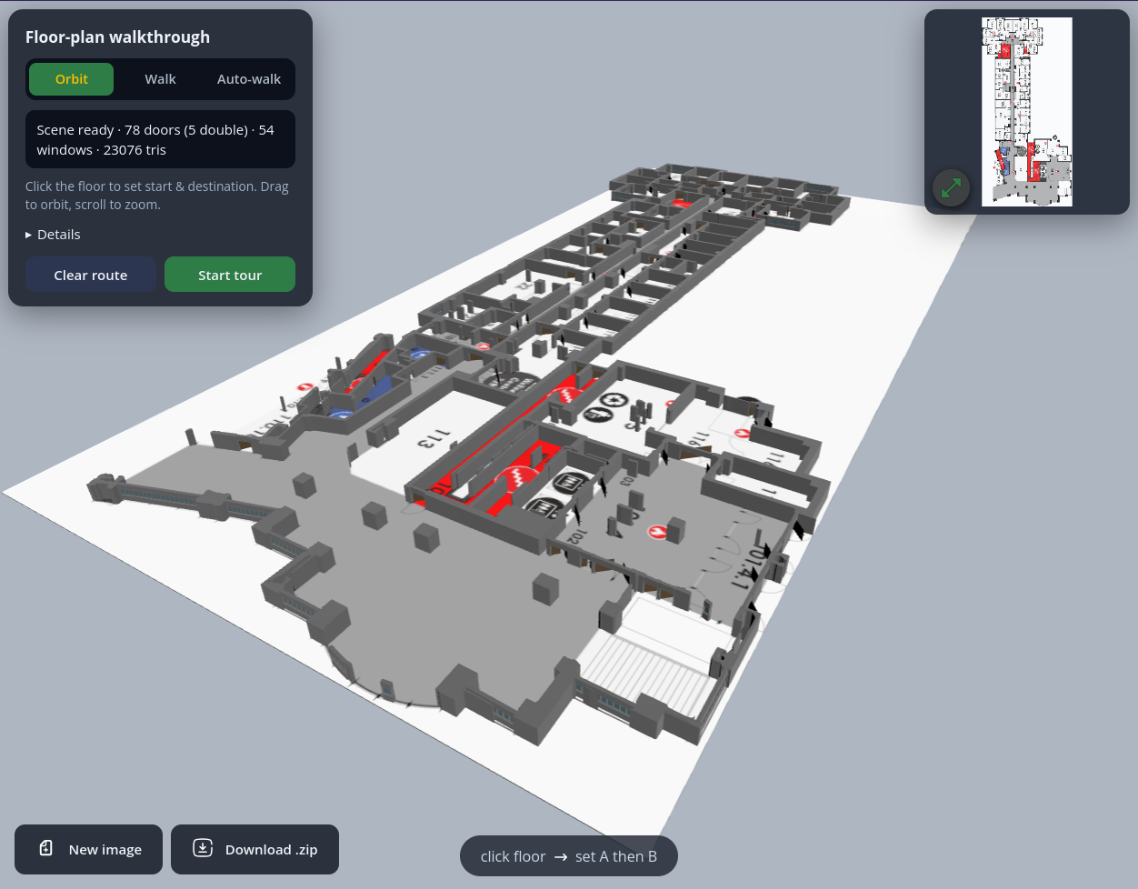}
\caption{The application's orbit view gives users an overview of 2D and 3D maps, with functionality such as map rotation and zooming}
\label{fig:gui-orbit}
\end{figure}

\begin{figure}[h!]
\centering    
\includegraphics[width=0.78\linewidth]{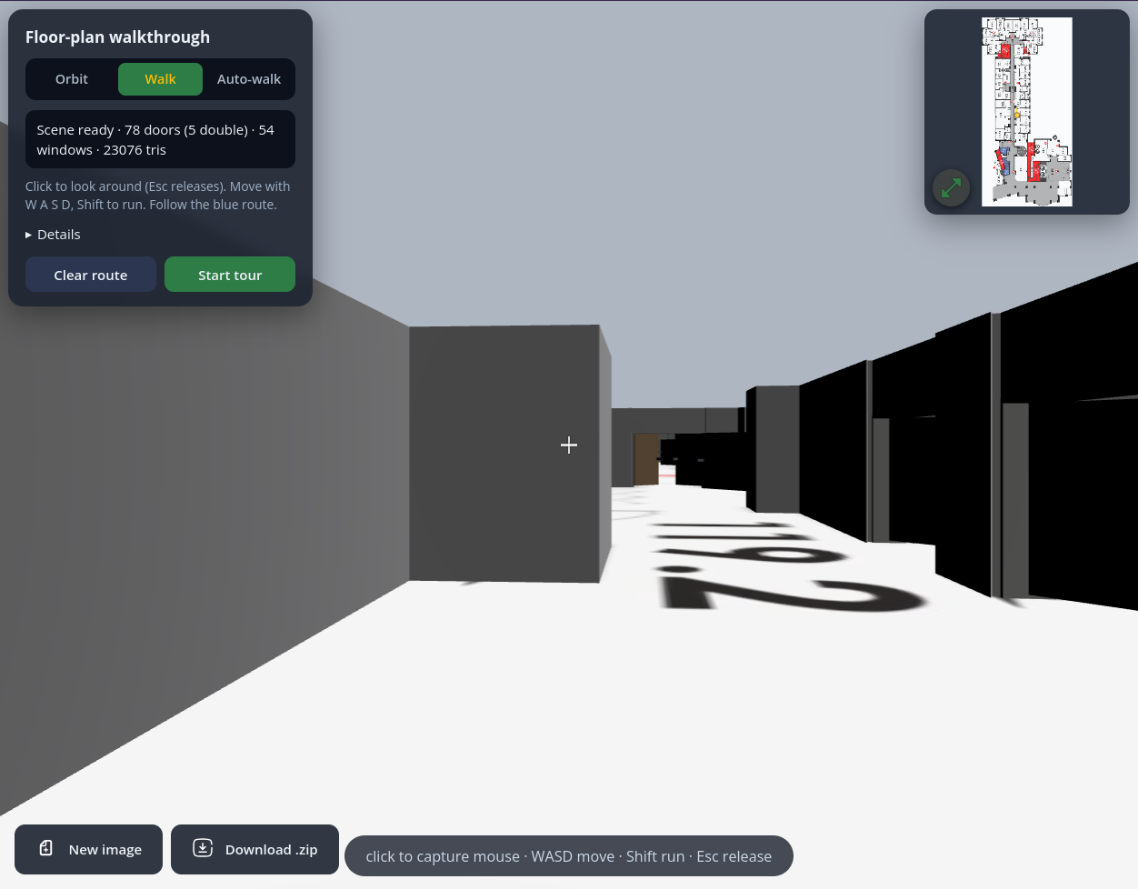}
\caption{The application has a walk view that is similar to the first-person view in a video game and users can explore the 3D map}
\label{fig:gui-walk}
\end{figure}

\begin{figure}[h!]
\centering    
\includegraphics[width=0.78\linewidth]{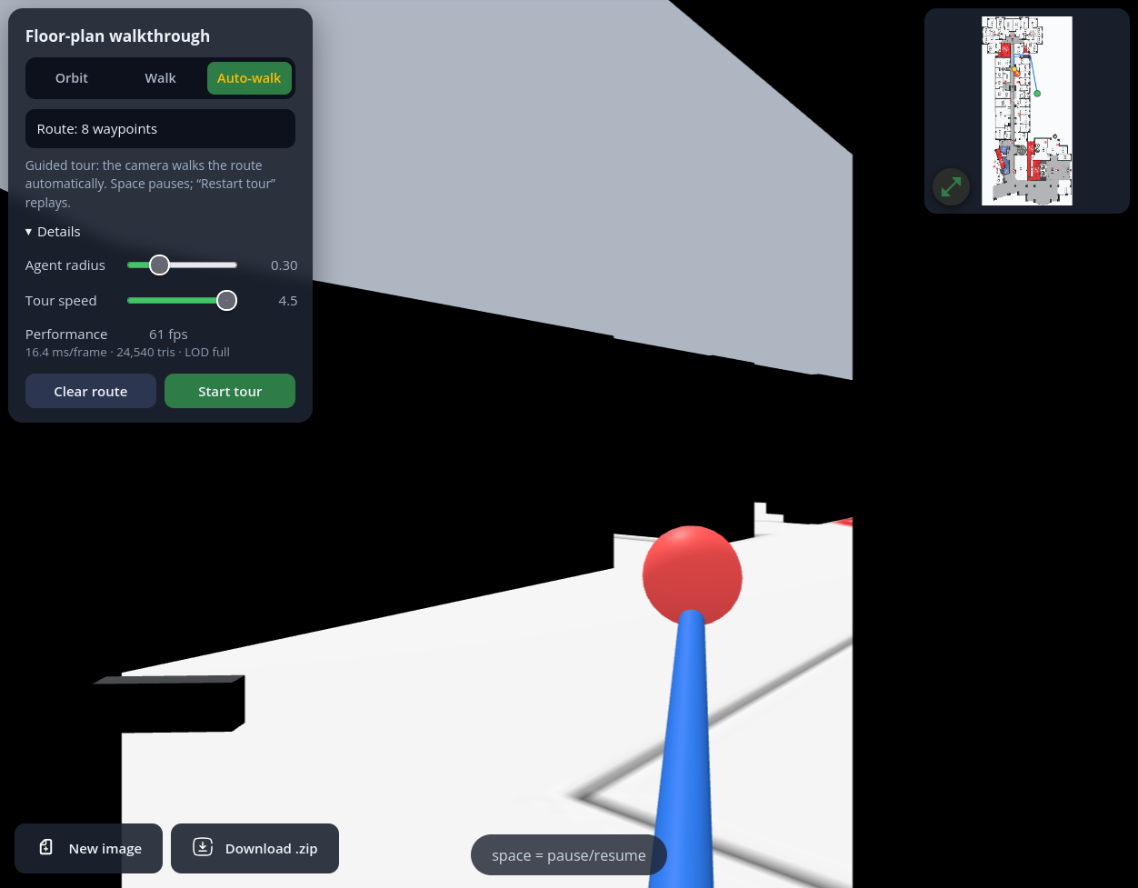}
\caption{The application provides an auto-walk view of the route, which the navigation algorithm uses to find the shortest path between the start and end points and the view automatically walks through the route from start to finish}
\label{fig:gui-autowalk_route}
\end{figure}

\begin{figure}[h!]
\centering    
\includegraphics[width=0.78\linewidth]{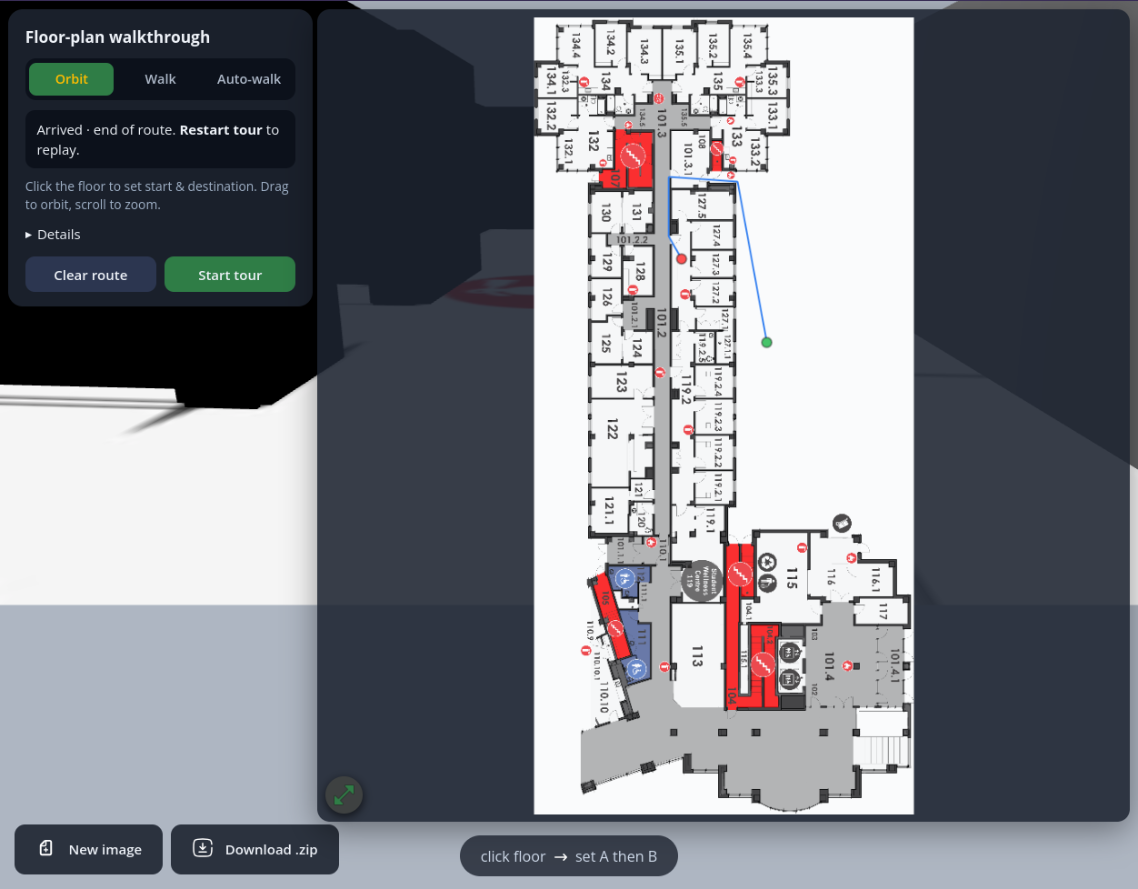}
\caption{Expanding the minimap provides a 2D map with all the details, as well as the routing path and the location of the view in walk and auto-walk modes}
\label{fig:gui-minimap}
\end{figure}

\section{GUI Testing}
Figures \ref{fig:gui-runtime} illustrates the home page of the simulator.
Figures \ref{fig:gui-orbit}, \ref{fig:gui-walk}, \ref{fig:gui-autowalk_route} and \ref{fig:gui-minimap} depict user interface in the Orbit, Walk, Auto-Walk and Minimap modes, alongside the navigation route.
The application's Orbit View (Figure \ref{fig:gui-orbit}) provides users with an overview of the map in both 2D and 3D, offering interactive features such as map rotation and zooming. 
The Walk View (Figure \ref{fig:gui-walk}) offers a first-person perspective, similar to that of a video game, allowing users to navigate and explore the 3D environment. 
The application also features an Auto-Walk View (Figure \ref{fig:gui-autowalk_route}), which provides an animated walkthrough of the calculated route. The navigation algorithm determines the shortest path between the specified start and end points, and the view automatically follows the route from beginning to end.
Additionally, expanding the Minimap displays a detailed 2D map (Figure \ref{fig:gui-minimap}), including all relevant map features and the calculated route. It also indicates the current position and viewing location in both Walk and Auto-Walk modes, allowing users to maintain spatial awareness while navigating the 3D environment.

\section{Conclusion and Future Work}
Our web-based AI-driven simulator transforms static indoor maps into interactive 3D environments with pathfinding and exploration. It aims to facilitate indoor navigation by enhancing users’ spatial awareness and helping them become familiar with large, complex and unfamiliar spaces. Our study presented an AI-based pipeline that transforms uploaded indoor plans into 2D and 3D representations. The proposed tool integrates synthetic data generation, semantic image segmentation, geometric reconstruction, wayfinding between specified locations, and interactive exploration. Its intended applications include efficient pathfinding, accessibility support for users with diverse mobility needs, and scenario-based navigation training in large and complex facilities.\\

Future work will focus on extending our tool into a gamified simulator that enables users to navigate between starting points and target destinations while encountering dynamically introduced obstacles and interactive game scenarios. The simulator will automatically generate navigation challenges and game rules to simulate real-world situations, offering varying levels of difficulty. Users will aim to reach designated destinations while avoiding obstacles, thereby enhancing their spatial awareness, navigation skills, and decision-making abilities in complex indoor environments.

\bibliographystyle{alpha}
\bibliography{Myreferences}
\end{document}